\documentclass{article}

\date{} % Certain latex templates automatically add the compilation date as a footnote in the generated pdf. This command can remove the date in some of the templates but does not work for all.

\usepackage{enumitem}

\usepackage{url}

\usepackage{hyperref} 
\usepackage{spconf,amsmath,epsfig,float,epstopdf}

\begin{document}\sloppy

%%%%%%%%%% for cover page
\onecolumn % make sure you keep this coverpage as one column. In this template, we force the coverpage to be one column with this command and then switch to double column for the remaining of the paper with the \doublecolumn command. 

\begin{description}[labelindent=1cm,leftmargin=4cm,style=multiline]

\item[\textbf{Citation}]{T. Alshawi, Z. Long, and G. AlRegib, ``Understanding spatial correlation in eye-fixation maps for visual attention in videos,'' Proceedings of IEEE International Conference on Multimedia and Expo (ICME), Seattle, WA, Nov. 2016.}
\\
\item[\textbf{DOI}]{\url{https://doi.org/10.1109/ICME.2016.7552913}}
\\
\item[\textbf{Review}]{Date of publication: 29 August 2016}
\\
\item[\textbf{Data and Codes}]{\url{https://ghassanalregibdotcom.files.wordpress.com/2016/10/tariq_icme2016_code.zip}}% If you do not have data related to this paper, you can remove the data keyword.
\\
\item[\textbf{Bib}] {@INPROCEEDINGS\{7552913, \\ 
author=\{T. Alshawi and Z. Long and G. AlRegib\}, \\ 
booktitle=\{2016 IEEE International Conference on Multimedia and Expo (ICME)\}, \\ 
title=\{Understanding spatial correlation in eye-fixation maps for visual attention in videos\}, \\ 
year=\{2016\}, \\ 
ISSN=\{1945-788X\}\}
} 
\\
% Preprint sharing policy can vary depending on the publisher. Before posting a paper to arXiv, please specifically check the transaction/conference you are targeting. In some transactions, papers are usually added to arxiv after acceptance. Publishers usually allow the authors to share accepted version of their papers but not the final formatted version that is provided by the publisher. In case of sharing preprints, publishers usually ask to add DOI and citation to the paper along with a copyright notice.

\item[\textbf{Copyright}]{\textcopyright 2016 IEEE. Personal use of this material is permitted. Permission from IEEE must be obtained for all other uses, in any current or future media, including reprinting/republishing this material for advertising or promotional purposes, creating new collective works, for resale or redistribution to servers or lists, or reuse of any copyrighted component of this work in other works.}
\\
\item[\textbf{Contact}]{\href{mailto:zhiling.long@gatech.edu}{zhiling.long@gatech.edu}  OR \href{mailto:alregib@gatech.edu}{alregib@gatech.edu}\\ \url{https://ghassanalregib.com/} \\ }
\end{description}

% Following command sequence was used to start the paper content from the following page and avoid numbering cover page.
\thispagestyle{empty}
\newpage
\clearpage
\setcounter{page}{1}

% Cover page was 1 column. \twocolumn changes the page format back to double column.
\twocolumn
%%%%%%%%%% for cover page

% Example definitions.
% --------------------
\def\x{{\mathbf x}}
\def\L{{\cal L}}

% Title.
% ------
\title{Understanding Spatial Correlation in Eye-fixation Maps for Visual Attention in Videos}
%
% Single address.
% ---------------
% \name{Anonymous ICME submission\vspace{20mm}}

% \address{}
\name{Tariq Alshawi, Zhiling Long, and Ghassan AlRegib}
\address{Center for Signal and Information Processing (CSIP)\\
School of Electrical and Computer Engineering\\
Georgia Institute of Technology, Atlanta, GA 30332-0250, USA\\
\{talshawi, zhiling.long, alregib\}@gatech.edu}

\maketitle

\begin{abstract}
In this paper, we present an analysis of recorded eye-fixation data from human subjects viewing video sequences. The purpose is to better understand visual attention for videos. Utilizing the eye-fixation data provided in the CRCNS (Collaborative Research in Computational Neuroscience) dataset, this paper focuses on the relation between the saliency of a pixel and that of its direct neighbors, without making any assumption about the structure of the eye-fixation maps. By employing some basic concepts from information theory, the analysis shows substantial correlation between the saliency of a pixel and the saliency of its neighborhood. The analysis also provides insights into the structure and dynamics of the eye-fixation maps, which can be very useful in  understanding video saliency and its applications.
% In this paper, we present an analysis of recorded eye-fixation data from human subjects viewing video sequences. The purpose is to better understand visual attention for videos. Utilizing the eye-fixation data provided in the CRCNS (Collaborative Research in Computational Neuroscience) dataset, this paper focuses on the relationship between the saliency of a pixel and that of its direct neighbors, without making any assumption about the structure of the eye-fixation maps. By employing some basic concepts from information theory, the analysis not only shows correlations between the saliency of a pixel and the saliency of its neighborhood in various scenarios, but also provides insights into the structure and dynamics of the eye-fixation maps. Further, our research provides an alternative quantitative approach to describing human attention, which can be very useful for various saliency-based applications.
\end{abstract}
\section{Introduction}
\label{sec:intro}
\nocite{Cover91}
Human visual attention modeling and understanding has been shown to be effective in analyzing big visual data as well as in improving the computation efficiency of visual data processing. Numerous applications have been proposed and currently investigated, such as object detection and recognition \cite{Ren2014saliencyRecognition}, scene understanding \cite{Bharath2013}, and multimedia summarization \cite{peng2010keyframe}.

\par To understand the visual attention mechanism, research usually relies on eye-tracking data analysis to formulate eye fixation maps. Such maps capture the focus of human subjects watching the videos and potentially correlate well with their visual attention. These maps are often used as the ground truth for saliency in learning-based methods, or as feature space for unsupervised methods. However, there has been limited research in the video processing community on analyzing these eye-fixation maps separately from saliency models. By studying the eye-fixation maps, we hope to better understand the spatial correlation in video scenes, and henceforth to better understand visual attention mechanisms.

\par The authors in~\cite{Sharma2014} analyzed eye-fixation data of images given location and time sequence of human subjects gaze, using the Eigen value decomposition of the correlation matrix constructed based on eye fixation data of different subjects. Their work shows that the first Eigen vector is responsible for roughly 21\% of the data, and it correlates well with salient locations in the images dataset. In \cite{Borji2013}, the authors found it is possible to decode the stimulus category by analyzing statistics (location, duration, orientation, and slope histograms) of fixations and saccades. They used a subset of the NUSEF dataset~\cite{Subramanian2010} containing five categories over a total of 409 images.

\par
In this paper, we analyze eye-fixation maps associated with spatiotemporal visual cues from the CRCNS~\cite{Itti05} dataset. In particular, we focus on spatial correlation in saliency. We investigate the possibility of predicting a pixel's saliency, as indicated by eye fixations, given the average saliency of its neighborhood. Since spatial correlation in the context of scene understanding refers to the relative locations of objects in the space, for videos it can be examined as the correlation between a pixel and its immediate spatiotemporal neighbors, as we will present in this paper. The contribution of this paper is two fold. First, we analyze the eye-fixation maps for videos, which has not been done in the literature, to our best knowledge. Second, contrary to other studies that foucs on general statstics of the eye-fixation map as a whole, we focus on the dynamics of a pixel with respect to its neighborhood in the eye-fixation map. Our research provides an alternative quantitative approach to describing human attention, which can be very useful for various saliency-based applications.

\section{Models}
\label{sec:models}
\subsection{Overall Map}
\label{subs:overall}
Given an eye-fixation map $F$ of size $M \times N$ and depth $K$ frames, first of all, we consider every map pixel $x[m,n,k] \in F$ to be an instance of a discrete integer random variable $X$ with an unknown distribution, that is:
\begin{equation}
\label{eq:variable}
X:\Omega \rightarrow E,
\end{equation}
where $\Omega$ is the set of all possible outcomes that describes the eye-fixation events during the experiments, $E$ is the observed set, and $x[m,n,k] \in \{0, 1, 2, ..., L\} $ numerically represents these events using $L+1$ symbols. For such a eye-fixation map $F$, we compute the Shannon entropy of $X$ as follows:
\begin{equation}
\label{eq:entropy}
H(X) = E[I(X)] = - \sum_{i=0}^L P(x_i)\log_2 P(x_i),
\end{equation}
where $E[\cdot]$ is the operator for expectation, $I(X)$ is the self-information of $X$, and $P(x_i)$ is the probability mass function for $X$.

\subsection{All Neighbors}
\label{subs:neighbors}

\begin{figure}
\centering
\includegraphics[width=8.5cm]{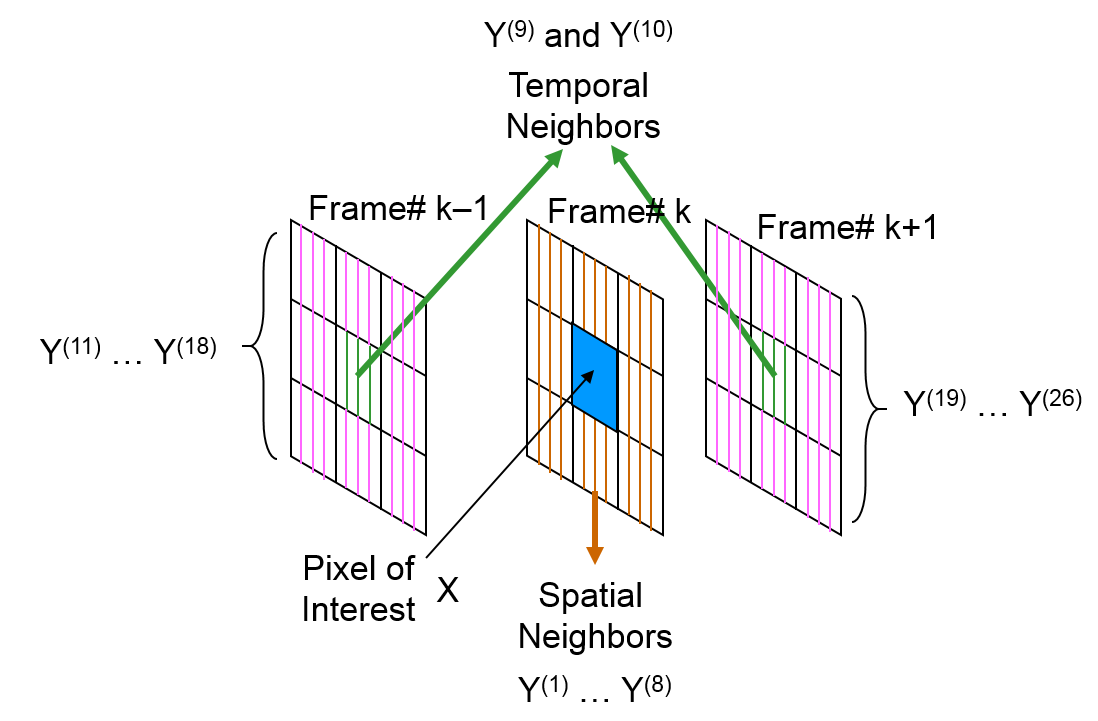}
\caption{Illustration of neighborhood pixels grouping. Spatial neighbors $Y^{(1)} ... Y^{(8)}$ of pixel $X$ are hashed in orange color, temporal neighbors $Y^{(9)}$ and $Y^{(10)}$ of $X$ are shown in green color, and the rest of spatiotemporal neighbors $Y^{(11)} ... Y^{(26)}$ are shown in purple.}
\label{fig:neighbors}
\end{figure}

For each map pixel, it is of interest to examine the relationship between the pixel and its neighbors. There are altogether 26 direct spatiotemporal neighbors (i.e., 9 pixels from frame $k-1$, 9 pixels from frame $k+1$, and 8 pixels from the current frame $k$), as shown in Fig.\ref{fig:neighbors}. To distinguish these neighbors from the center pixel, we label them as $Y^{(j)}$, where $j \in \{1, 2, ..., 26\}$. We compute the conditional entropy of a center pixel $X$ given the average of its direct neighbors as follows:
\begin{equation}
\label{eq:conditionalEntropy}
H(X|Z) = \sum_{\forall x_i,z_j}  P(x_i,z_j) \log_2 \frac{P(z_i)}{P(x_i,z_j)}
\end{equation}
where $Z = f\big(Y^{(1)},\dots, Y^{(26)}\big)$ is the arithmetic mean of the 26 direct neighbors, and $P(x_i,z_j)$ is the joint probability mass function for $X$ (the center pixel) and $Z$ (the mean of its direct neighbors). As a basic property of conditional entropy, the following relationship always holds:
\begin{equation}
\label{eq:property}
0 \le H(X|Z) \le H(X).
\end{equation}
Here, if $Z$ completely determines $X$, then $H(X|Z) = 0$; or, if $X$ is completely independent of $Z$, then $H(X|Z) = H(X)$, which means knowing $Z$ does not reduce the uncertainty about $X$.

\subsection{Spatial Neighbors}
\label{subs:spatial}
In addition to analyzing the correlation between a pixel and its neighbors in the general sense, we investigate the effect of the video content on such correlation. To do this, we need to extend the model introduced above. Now we consider all pixels at location $[m,n]$ in the eye-fixation map across all $K$ frames as instances of a random variable $X[m,n]$. Similarly, we calculate the arithmetic mean, denoted as $Q[m,n]$, of the eight direct spatial neighbors $X[m+i,n+j]$, where $i$ and $j \in \{ 1,0,-1\}$, as shown in Fig.\ref{fig:neighbors}. To quantify their correlation, we compute mutual information between $X[m,n]$ and $Q[m,n]$ as follows:
\begin{equation}
\label{eq:mutualInformation}
I(X[m,n];Q[m,n]) = \sum_{\forall x_i,q_j}  P(x_i,q_j) \log_2 \frac{P(x_i,q_j)}{P(x_i)P(q_j)},
\end{equation}
where $P(x_i)$ is the probability mass function of random variable $X[m,n]$, $P(q_j)$ is the probability mass function of $Q[m,n]$, the arithmetic mean of the spatial neighbors, and $P(x_i,q_j)$ is the joint probability mass function of $X[m,n]$ and $Q[m,n]$.

\subsection{Temporal Neighbors}
\label{subs:temporal}
Similar to the correlation with spatial neighbors, we analyze the correlation with temporal neighbors. For this purpose, we modify the model again as follows. First, we consider each pixel in frame $k$ of an eye-fixation map, $F(k)$, as an instance of a random variable $X_k$. Then, we obtain another random variable $W_k$, which represents a pixel-wise arithmetic mean of adjacent frames $F(k+D)$. Here, $D \in \{\pm1,\pm2,\pm3, \dots\}$. Similar to Sec.\ref{subs:spatial}, we quantify the eye-fixation correlation with temporal neighbors by computing the mutual information between $X_k$ and $W_k$ as follows:
\begin{equation}
\label{eq:mutualInformationT}
I(X_k;W_k) = \sum_{\forall x_i,w_j}  P(x_i,w_j) \log_2 \frac{P(x_i,w_j)}{P(x_i)P(w_j)},
\end{equation}
where $P(x_i)$ is the probability mass function of $X_k$, $P(w_j)$ is the probability mass function of $W_k$, and $P(x_i,w_j)$ is the associated joint probability mass function.

\begin{figure*}[t]
\centering
\includegraphics[width=0.9\textwidth]{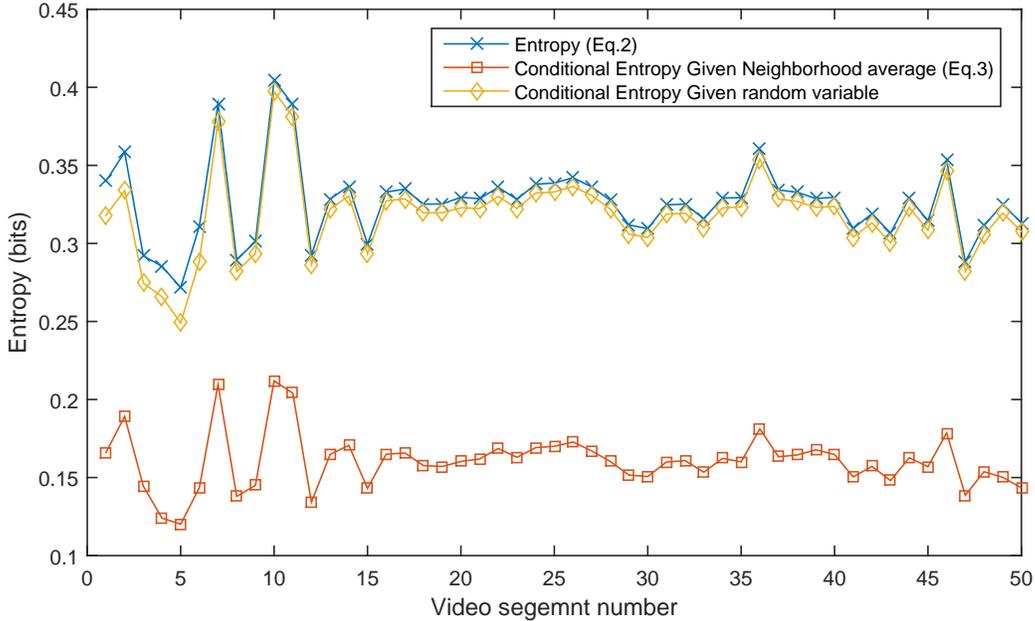}
\caption{Entropy reduction over all video segments.}
\label{fig:EntropyReduction}
\end{figure*}

\section{Experiments}
\label{sec:exp}

\subsection{Preparing Eye-fixation Data}
\label{sec:data}
% \noindent
Our study in this paper uses eye-fixation maps from the public CRCNS database \cite{Itti05}. The database includes 50 videos, with a resolution of $480 \times 640$, durations ranging between $5$ and $90$ seconds, and a frame rate of $30$fps. The videos are diverse with a total of $12$ categories ranging from street scenes to video games and from TV sports to TV news. In many cases the videos contain variations of lighting conditions, severe camera movements, and high motion blur effects. The eye tracking data were collected from eight subjects using an ISCAN RK-464 eye-tracker at 240 $Hz$ sampling rate, which was calibrated every five clips using 9-point calibration. The stimuli were displayed on $22"$ CRT monitor at $~80cm$ viewing distance with mean screen luminance of 30 $cd/m^2$. Eye tracking data are provided for each human subject separately in a string of eye gaze coordinates, which span 0 to 639 in the horizontal direction and 0 to 479 in the vertical direction with location (0,0) being at the top left corner of the monitor. Labels are available for each eye-gaze sample, e.g., fixation, saccade, and during blink, just to name a few.

For a given video sequence, we prepare an eye fixation map from the eye-tracking data in the following manner:
% (Fig.\ref{fig:procedure} depicts the procedure):
\begin{enumerate}
  \item Construct a frame of size 480 $\times$ 640 and initialize it to zeros.
  \item Collect all eye-gaze samples corresponding to a given video frame. We only select samples that are fixation or smooth pursuit. Saccade or loss-of-tracking samples are not included.
  \item For every sample obtained in step two, we set the pixel value at the corresponding location to one. If two samples coincide in spatial location, we set the pixel value equal to the number of samples pointing to that location.
  \item Following the procedure above, we process the eye-tracking data frame by frame, and finally construct an eye-fixation map with the same size and number of frames as the video sequence.
\end{enumerate}

% \begin{figure}
% \centering
% \includegraphics[width=6.5cm]{Figures/Procedure2.PNG}
% \caption{Illustration of the procedure for preparing the eye-fixation data.}
% \label{fig:procedure}
% \end{figure}

Additionally, we construct the eye-fixation maps at various scales by reducing the original map size, which are useful in practical applications when video frames are often processed at reduced frame sizes. For an original map $F(m,n,k)$, the size-reduced map of scale $s$, $F^{(s)}(m,n,k)$, is formed as
\begin{equation}
\label{eq:scaling}
F^{(s)}(m,n,k) = \sum_{\forall i,j \in R_s}  F(i,j,k)
\end{equation}
where $R_s$ is the window that contains pixels of $F(m,n,k)$ corresponding to pixel $(m,n,k)$ in $F^{(s)}(m,n,k)$.

\subsection{Results and Discussions}
\label{sec:results}

\subsubsection{Correlation with All Neighbors}
In the following experiments, we set $R_s$ to 40 $\times$ 40 window size. This helps reduce the computation time and still generates results similar to those obtained using the original map size. First, we evaluate the correlation between a map pixel and its direct spatiotemporal neighbors as detailed in Sec.~\ref{sec:models}. We plot the entropy values computed for each of the 50 video sequences in the dataset in Fig.\ref{fig:EntropyReduction}. As shown in the figure, the entropy of the eye-fixation drops when the spatiotemporal neighborhood average is considered, which is the red curve in the plot. To have a basis for comparison, we also calculate an entropy conditioned on a uniformly-distributed random variable and show the results as the yellow curve in the same figure.

The reduction in the entropy values in many cases reaches ~50\% in the red curve. Such drop value is significant when compared to reduction in entropy due to conditioning on the uniformly-distributed random variable, indicating a strong and meaningful correlation. Another important observation from Fig.\ref{fig:EntropyReduction} is that the entropy reduction is consistent across all videos in the dataset. The average entropy reduction is 0.0815 bits with variance $3.2416\times10^{-05}$.

It is worth noting that the entropy of the eye-fixation maps is generally low due to the sparsity of such maps. In fact, most of the map pixels have a value that equals to zero. However, since the probability mass is concentrated in a single symbol, the skewness of the probability mass function does not affect statistical test outcome. It merely moves the entropy (and the conditional entropy in turn) up or down. The three video segments with the highest entropy values are \textsf{gamecube02}, \textsf{gamecube06}, and \textsf{gamecube13}. These videos have relatively longer duration and engaging content, which potentially enable more cognitive processes to take place, thus contributing to the higher entropy.

\subsubsection{Correlation with Spatial Neighbors}
\par
\begin{figure}
\centering
\begin{tabular}{c}
\includegraphics[width=0.3\textwidth]{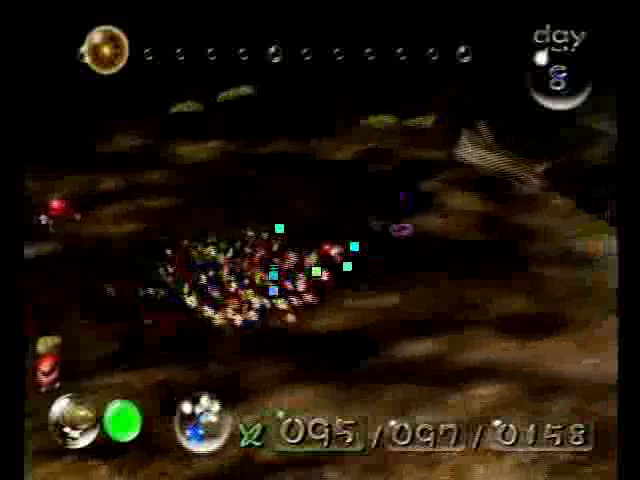} \\
(a) \textsf{gamecube06} sample frame taken at \textsf{01m:57s:076'} \\
\includegraphics[width=0.35\textwidth]{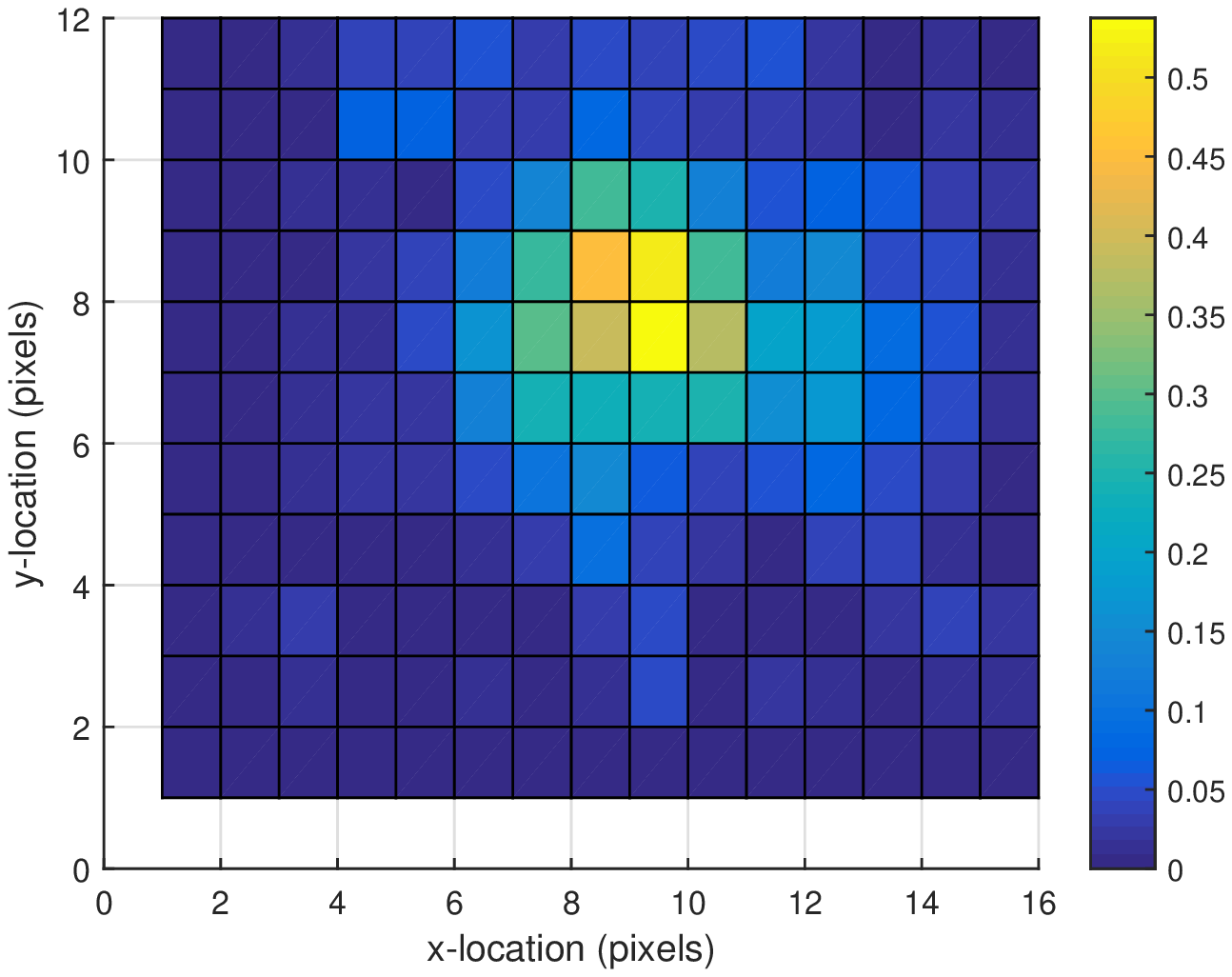}\\
(b) \textsf{gamecube06} mutual information given spatial location.\\
\end{tabular}
\caption{\textsf{gamecube06} sample frame alone with mutual information given spatial location, involving only spatial neighbors}
\label{fig:gamecube}
\end{figure}

 Second, we study the correlation between a map pixel in a given spatial location and its direct spatial neighborhood. The purpose is to evaluate the impact of the famous center-bias phenomenon \cite{Goa2008} on the correlation. It should be noted here that the researchers collecting the database have every video segment preceded by a blinking cross in the middle of the screen, exactly at [239,319]. The blinking lasts for  1 $sec$ before the video is shown. Consequently, \emph{lack of knowledge} center-bias is present in almost all videos. However, it contributes mostly to the viewing of the first few frames, thus having not much effect on the overall correlation. The majority of the videos in this dataset have another center-bias factor that significantly influences the end results, i.e., the \emph{photography} center-bias. This bias is due to the tendency of photographers to place the object(s) of interest at the center of the video frames. Even though many image datasets for visual attention research have taken this into account, it is difficult to do the same for videos. This photography center-bias is particularly obvious in the \textsf{gamecube} video segments, with a sample frame shown in Fig.\ref{fig:gamecube}.(a), since the in-game camera system is designed to place the game character(s) in the center of the video.

As described in Sec.\ref{subs:spatial}, we compute the mutual information between a pixel at a given location and its direct spatial neighbors. The results are shown in Fig.\ref{fig:gamecube}.(b). We can clearly notice a high correlation in the case of \textsf{gamecube06}, as exists in virtually every \textsf{gamecube} video, which can be attributed to the photography center-bias. It is surprising to notice that even though textual information is located at the corners of the screen, it does not attract eye-fixation for prolonged periods due the relatively low information content they convey.

\begin{figure}
\centering
\begin{tabular}{c}
\includegraphics[width=0.3\textwidth]{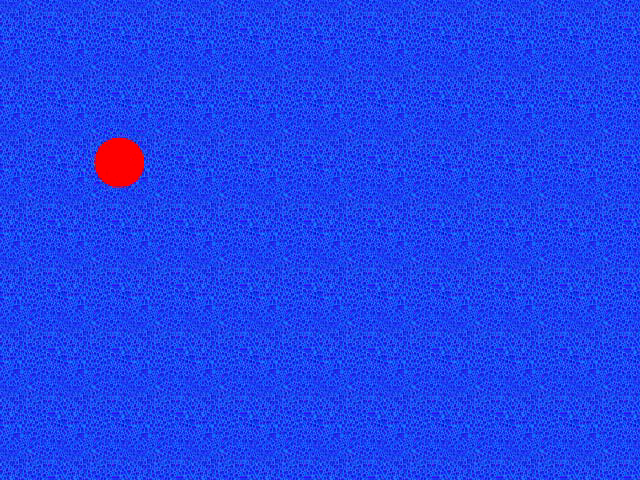}\\
(a) \textsf{saccadetest} sample frame taken at \textsf{00m:07s:606'}.\\
\includegraphics[width=0.35\textwidth]{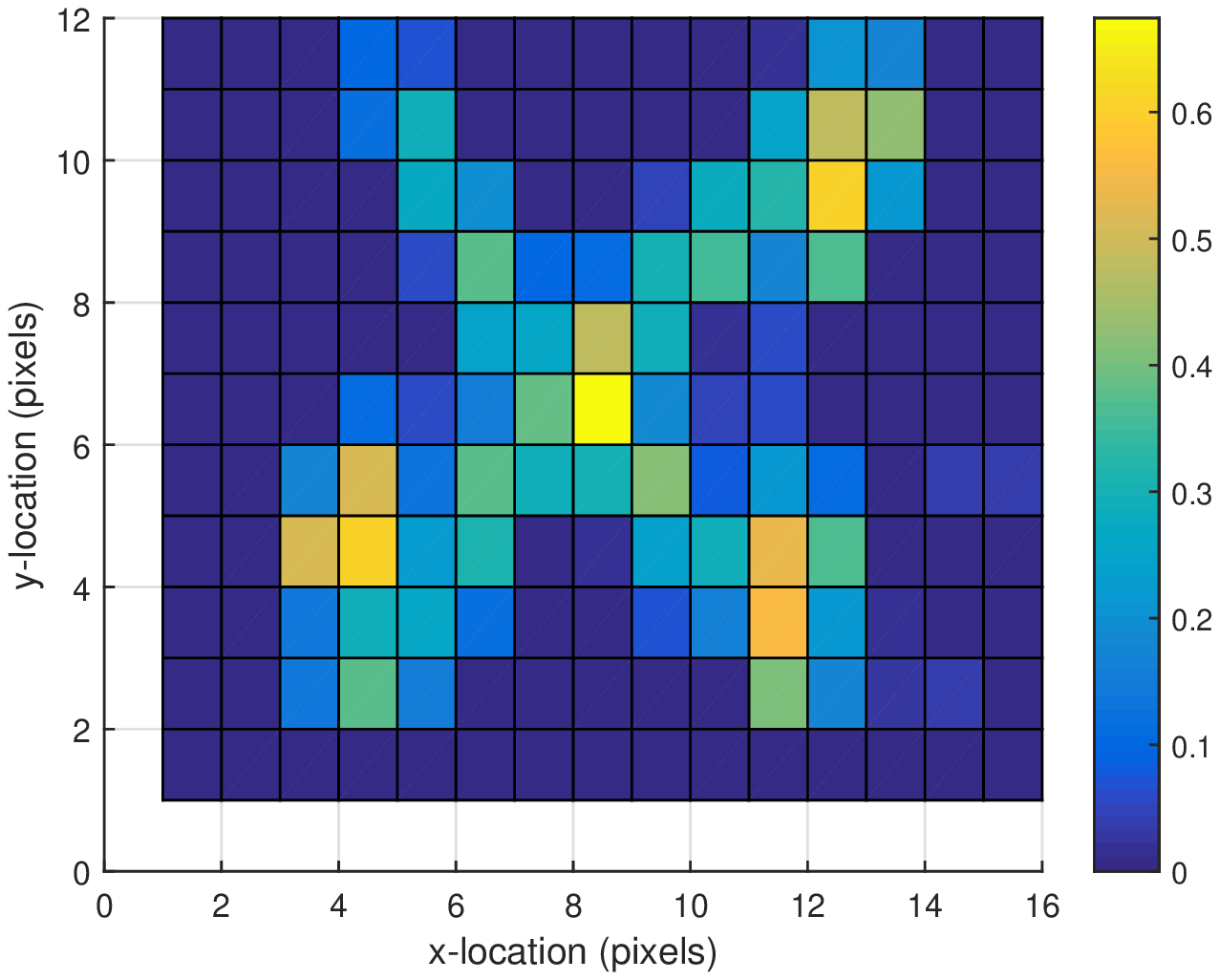}\\
(b) \textsf{saccadetest} mutual information given spatial location.\\
\end{tabular}
\caption{\textsf{saccadetest} sample frame alone with mutual information given spatial location}
\label{fig:saccadetest}
\end{figure}

On the other hand, videos that lack photography center-bias exhibit totally different behaviour. For example, the \textsf{saccadetest} video, with a sample frame shown in Fig.\ref{fig:saccadetest}.(a), consists of a red dot against a blue textured background. The dot is static and changes places for the first half of the clip, then it moves diagonally with a consistent speed for the rest of the video. As shown in Fig.\ref{fig:saccadetest}.(b), the correlation is highest when there is a smooth pursuit following the red dot, due to high sampling rate and low spatial displacement in the object of interest. Similar trends can also be observed in video segments such as \textsf{beverly06}, \textsf{beverly07}, and \textsf{beverly08}.

\par
\begin{figure}
\centering
\begin{tabular}{c}
\includegraphics[width=0.3\textwidth]{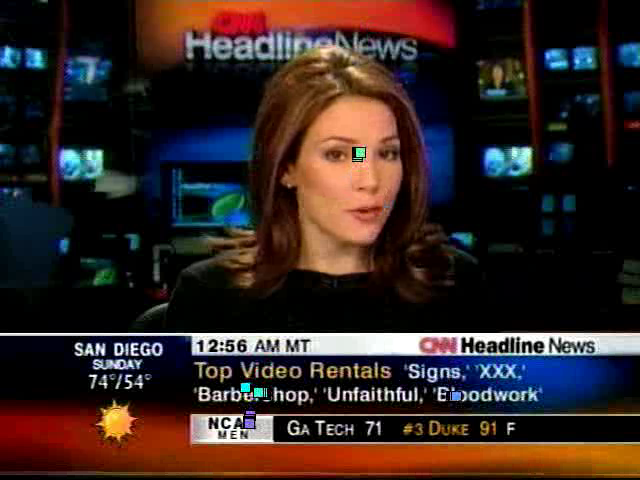}\\
(a) \textsf{tv-news03} sample frame taken at \textsf{06m:51s:026'}.\\
\includegraphics[width=0.35\textwidth]{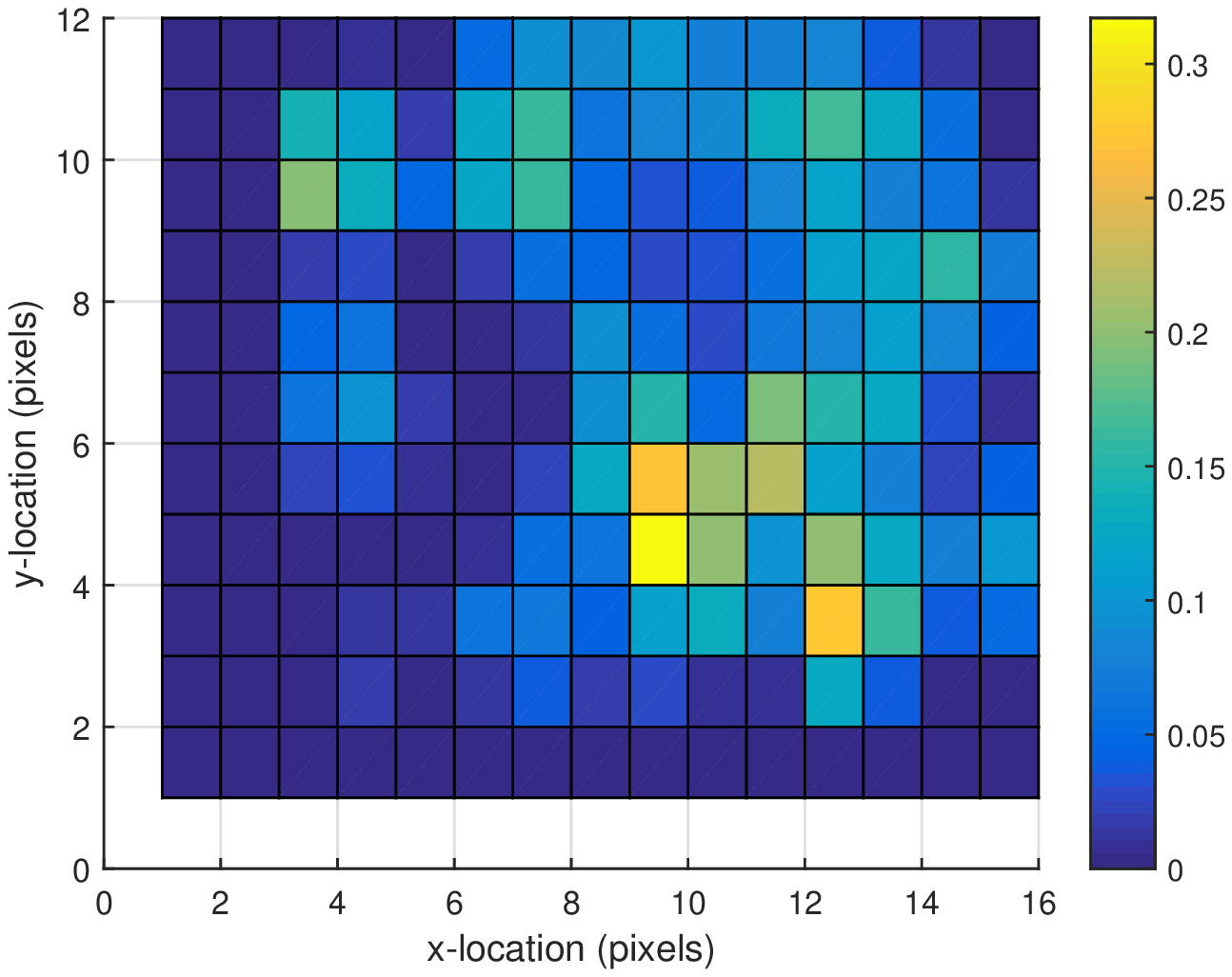}\\
(b) \textsf{tv-news03} mutual information given spatial location.\\
\end{tabular}
\caption{\textsf{tv-news03} sample frame alone with mutual information given spatial location}
\label{fig:tv_news03}
\end{figure}

 Additionally, interesting trends can be observed when there are multiple salient objects present in the scene, such as in the \textsf{tv-news03} video, a sample frame of which shown in Fig.\ref{fig:tv_news03}.(a). The space-localized mutual information map, shown in Fig.\ref{fig:tv_news03}.(b), exhibits two centers of attention. One corresponds to the most semantically informative object in the scene, i.e., the news anchor's face. The other is the textual messages in the lower banner. Since the human subjects spend considerable periods of time looking at these two locations, the correlation is significantly higher than other locations in the eye-fixation map. The example results, especially those from the latter two without obvious center-bias, demonstrate that the higher correlation areas match very well with the human attention.

\subsubsection{Correlation with Temporal Neighbors}
\par
\begin{figure}
\centering
\includegraphics[width=0.48\textwidth]{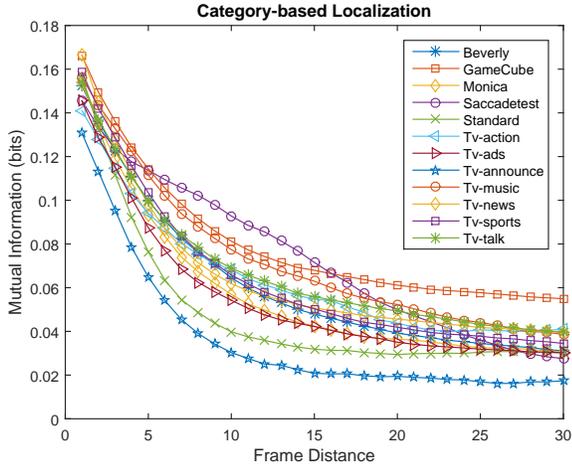}
\caption{Average mutual information between $F(k)$ and temporal neighbors $F(k+D)$ and $F(k-D)$, where $D$ is the frame distance.}
\label{fig:temporalCorrelationLocalized}
\end{figure}

 Finally, we study the correlation over time as described in Sec.\ref{subs:temporal}. We begin by computing mutual information between a given frame $F(k)$ and the average of its temporal neighbors $F(k+D)$ and $F(k-D)$ with different values of distance $D$. As shown in Fig.\ref{fig:temporalCorrelationLocalized}, mutual information between a frame and its direct neighbors (i.e., $D=1$) is significant compared to the information shared with distant frames. For all videos, roughly 50\% of information is shared between adjacent neighboring frames (recall that the average information content in an eye-fixation map is about 0.3 bits as shown in Fig.\ref{fig:EntropyReduction}). Another important realization is that this trend is seen in every category in the dataset, regardless of the content. However, the rate of change in some categories are small, such as \textsf{saccadetest}, and some are large, such as \textsf{tv-ads} and \textsf{tv-sport}. These differences can be explained by the level of complexity of these videos. Simple videos (e.g., \textsf{saccadetest}) have very few stimuli. Thus, human subjects tend to fixate on a single target, which results in more correlated frames in the eye-fixation map. On the other hand, complex videos (e.g., \textsf{tv-ads} and \textsf{tv-sports}) contain much more stimuli, requiring more efforts from the human subjects to examine and comprehend the scene.

\begin{figure}[h]
\centering
\includegraphics[width=0.48\textwidth]{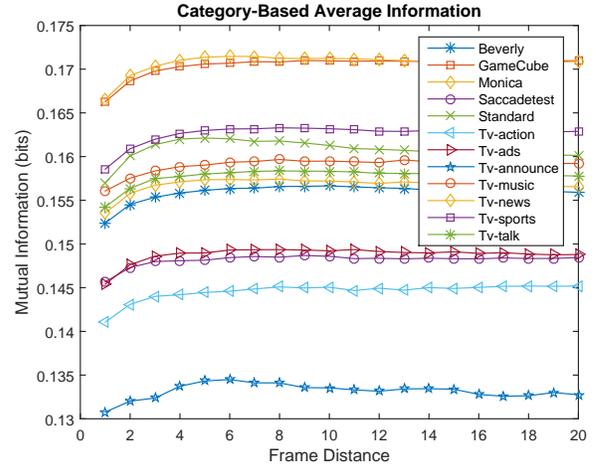}
\caption{Average mutual information between $F(k)$ and the average of its $N$ temporal neighbors up to a certain frame distance.}
\label{fig:temporalCorrelationAverage}
\end{figure}

Moreover, we compute the correlation between a given frame $F(k)$ and the average of its $N$ direct neighbors up to a certain frame distance. As shown in Fig.\ref{fig:temporalCorrelationAverage}, mutual information between a given frame and its nearest neighbors contains most of the information shared with the rest of the frames. For most categories, the nearest $5-6$ neighbors contain almost all the correlated information in the eye-fixation map. Therefore, including more frames in the neighborhood average does not necessarily add any more useful information. This trend is observed in every category in the dataset, regardless of the content. However, some categories (such as \textsf{monica} and \textsf{gamecube}) yield higher mutual information than other categories, suggesting that the video content makes some differences. For most categories, the mutual information level-off after 8-10 frames, with the exception of \textsf{standard}. This can be attributed to the process of averaging that may cause some mutual information in the nearest neighbors to be marginalized as the number of frames included gets greater.

\section{Conclusions}
\label{sec:conclusions}
In this paper, we presented an information-theory-based analysis of recorded eye-fixation data from human subjects viewing video sequences to gain insights into visual attention mechanisms for videos. The analysis focused on the relationship between the saliency of an eye-fixation map pixel and that of its neighbors. Our experiments demonstrated that a substantial correlation between the saliency of a pixel and the saliency of its neighborhood exist. Such correlation is localized both spatially and temporally, and is significantly affected by the video's content and complexity. Our research provides an alternative quantitative approach to describing human attention. We believe such an approach is very important for many saliency applications. For example, ground truth data for saliency detection can be changed from the traditional eye-fixation data into a more descriptive format based on the correlations. The various correlations discussed in the paper can also be used as measures of the reliability of detected saliency, thus being a guide for optimizing saliency-based video processing.

\bibliographystyle{IEEEbib}
\bibliography{main}

\end{document}